%% file: final.tex
\documentclass[10pt,twocolumn,letterpaper]{article}
\usepackage{amsmath}
\usepackage{cases}
\usepackage{cvpr}
\usepackage{times}
\usepackage{subcaption}
\usepackage{epsfig}
\usepackage{graphicx}
\usepackage{amsmath}
\usepackage{amssymb}
\usepackage{multirow}
\usepackage{multicol}
\usepackage[breaklinks=true,bookmarks=false]{hyperref}

\usepackage{times}
\usepackage{epsfig}
\usepackage{graphicx}
\usepackage{amsmath}
\usepackage{amssymb}

\usepackage{algorithm}
\usepackage{algorithmicx}
\usepackage{algpseudocode}
\usepackage{color}
\usepackage{enumitem}
\usepackage{arydshln}
\usepackage{multirow}
\usepackage{bm}

\def\l1{{$l_1$}}
\def\L1{{$l_1$}}

\usepackage[margin=4pt,font=footnotesize,labelfont=bf,labelsep=endash,tableposition=top]{caption}

\renewcommand{\texttt}[1]{ $ {{\tt #1} } $}

\cvprfinalcopy 

\def\cvprPaperID{5639} % *** Enter the CVPR Paper ID here

\begin{document}

\title{PolarMask: Single Shot Instance Segmentation with Polar Representation}

\author{
{\large
Enze Xie$^{1,2*}$, ~Peize Sun$^{3}$\thanks{ indicates equal contribution.}, ~Xiaoge Song$^{4*}$, ~Wenhai Wang$^{4}$, }\\
{\large
Ding Liang$^{2}$, ~Chunhua Shen$^{5}$, ~Ping Luo$^{1}$}
\\[.21cm]
{\large
${^1}$The University of Hong Kong ~~~~${^2}$Sensetime Group Ltd}\\
{\large
${^3}$Xi'an Jiaotong University ~~~~${^4}$Nanjing University ~~~${^5}$The University of Adelaide}\\
\small E-mail: $\tt xieenze@hku.hk $ 
}

\maketitle
\begin{abstract}

In this paper, we introduce an anchor-box free and single shot instance segmentation method, which is conceptually  simple, fully convolutional and can be used by easily embedding it into most off-the-shelf detection methods. 
Our method, termed PolarMask, formulates the instance segmentation problem as predicting contour of instance through instance center classification and dense distance regression in a polar coordinate. 
Moreover, we propose two effective approaches to deal with sampling high-quality center examples and optimization for dense distance regression, respectively, which can significantly improve the performance and simplify the training process.
Without any bells and whistles, PolarMask achieves 32.9\% in mask mAP with single-model and single-scale training/testing on the challenging COCO dataset.

For the first time, we show that the complexity of instance segmentation, in terms of both design and computation complexity, can be the same as bounding box object detection and this much simpler and flexible instance segmentation framework can achieve competitive accuracy. 
We hope that the proposed PolarMask framework can serve as a fundamental and strong baseline for single shot instance segmentation task. 
Code is available at:
\href{https://github.com/xieenze/PolarMask}{\color{blue}{$\tt github.com/xieenze/PolarMask$}}.
 
\end{abstract}

\section{Introduction}

\begin{figure}[ht]
\centering
\begin{minipage}[b]{0.23\textwidth}
  \centering
  \centerline{\includegraphics[width=25mm]{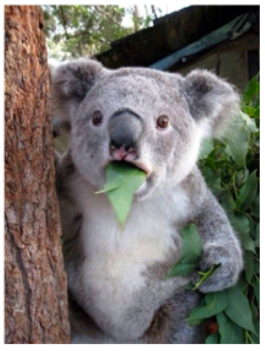}}
  \centerline{(a) Original image}
\end{minipage}
\hspace{0.2mm}
\begin{minipage}[b]{0.23\textwidth}
  \centering
  \centerline{\includegraphics[width=25mm]{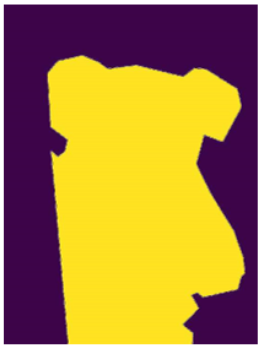}}
  \centerline{(b) Pixel-wise Representation}
\end{minipage}
\vspace{3mm}
\begin{minipage}[b]{0.23\textwidth}
  \centering
  \centerline{\includegraphics[width=45mm]{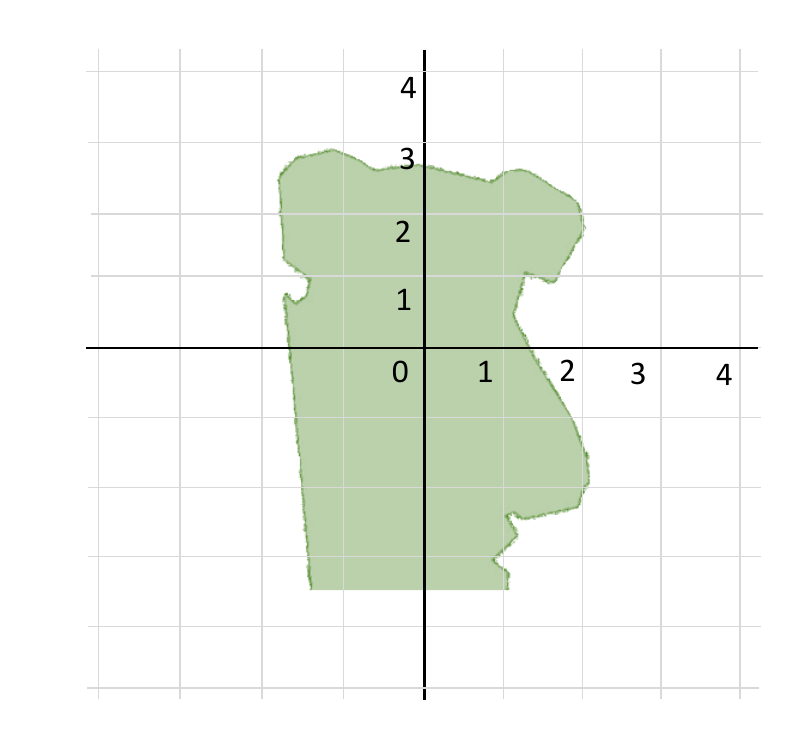}}
  \centerline{(c) Cartesian Representation}
\end{minipage}
\hspace{1mm}
\begin{minipage}[b]{0.23\textwidth}
  \centering
  \centerline{\includegraphics[width=45mm]{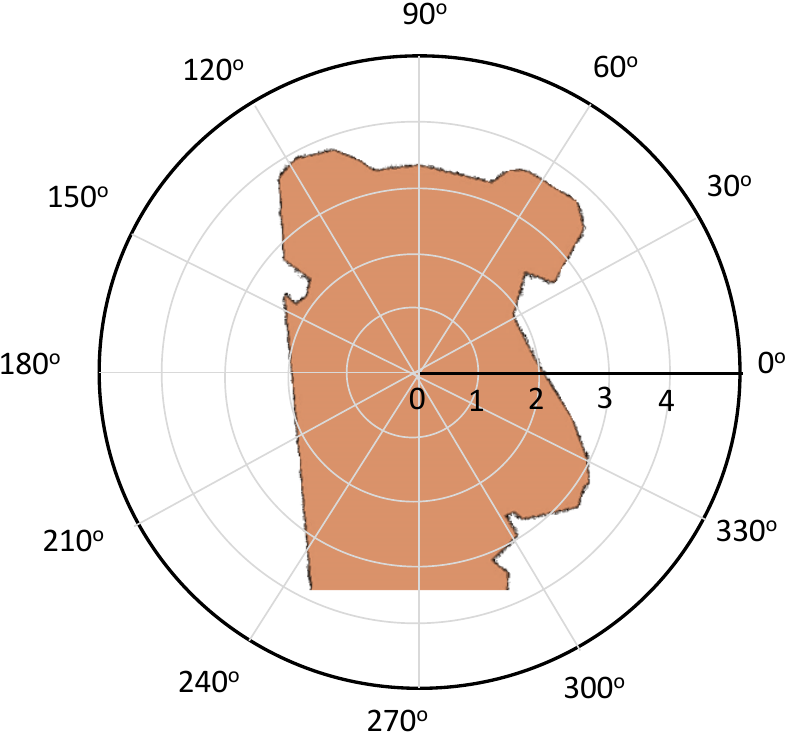}}
  \centerline{(d) Polar Representation}
\end{minipage}

\caption{Instance segmentation with different mask representations. (a) is the original image. (b) is the pixel-wise mask representation. (c) and (d) represent a mask by its contour, in the Cartesian and Polar coordinates, respectively.}
\label{fig:intro}
\end{figure}

Instance segmentation is one of the fundamental tasks in computer vision, which enables numerous downstream vision applications.
It is challenging as it requires  to predict both the location 
and the semantic mask of each instance in an image. Therefore intuitively  instance segmentation can be solved by bounding box detection then semantic segmentation within each box, adopted by two-stage methods, such as Mask R-CNN~\cite{maskrcnn}.    
Recent trends in the vision  community have spent more effort in designing simpler pipelines of bounding box detectors~\cite{densebox,focalloss,FCOS,reppoints,objectspoints,centernet,foveabox} and subsequent instance-wise recognition tasks including instance segmentation~\cite{yolact,tensormask,extremenet}, which is also the main focus of our work here. 
\textit{
    Thus, our aim is to design a conceptually simple mask prediction module that can be easily plugged into many off-the-shelf detectors, enabling instance segmentation.     
}

\begin{figure*}[!t]
\begin{center}
\includegraphics[width=0.99\textwidth]{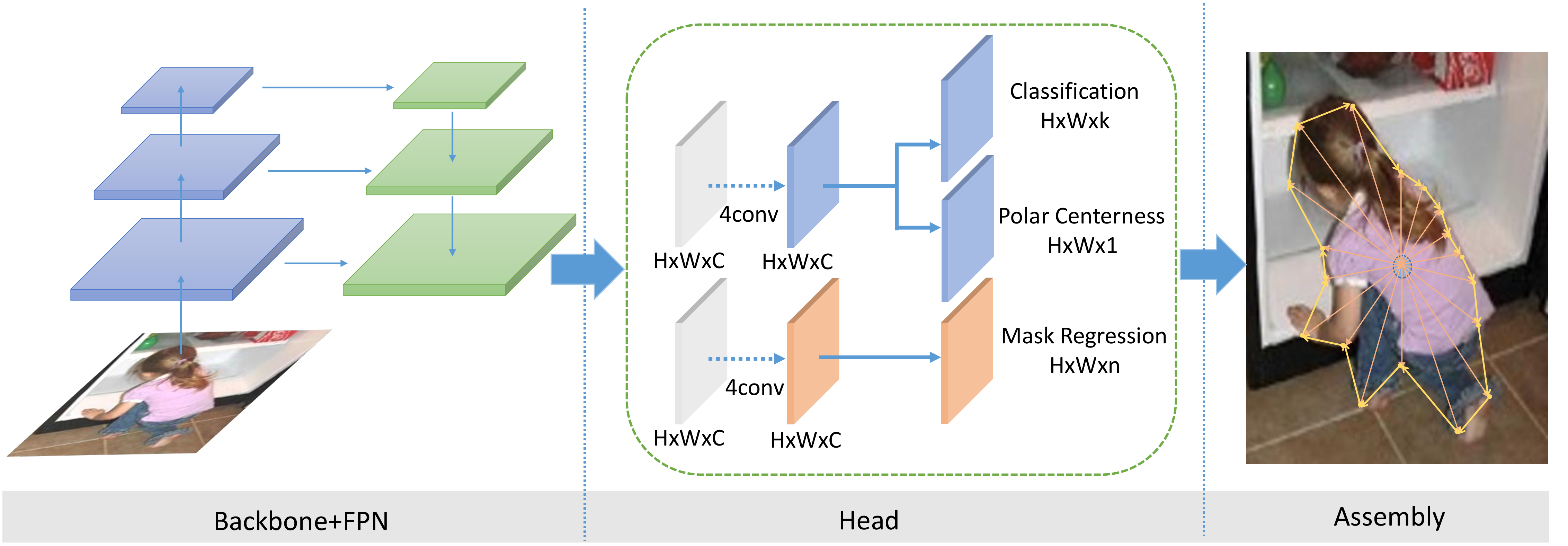}
\caption{The overall pipeline of PolarMask. The left part contains the backbone and feature pyramid to extract features of different levels. The middle part is the two heads for classification and polar mask regression. $H, W, C$ are the height, width, channels of feature maps, respectively, and $k$ is the number of categories (e.g., $k =80$ on the
COCO dataset), $n$ is the number of rays (e.g., $n=36$)}. 
\label{fig:pipeline}
\end{center}
\vspace{-3mm}
\end{figure*}

Instance segmentation is usually solved by binary classification in a spatial layout surrounded by bounding boxes, shown in Figure~\ref{fig:intro}(b). Such pixel-to-pixel correspondence prediction is luxurious, especially in the single-shot methods. Instead, we point out that masks can be recovered successfully and effectively if the contour is obtained. An intuitive method to locate contours is shown in Figure~\ref{fig:intro}(c), which predicts the Cartesian coordinates of the point composing the contour. Here we term it as Cartesian Representation. The second approach is Polar Representation, which applies the angle and the distance as the coordinate to locate points, shown in Figure~\ref{fig:intro}(d).

In this work, we design an instance segmentation  method based on the Polar Representation since its inherent advantages are as follows:
(1) The origin point of the polar coordinate can be seen as the center of object. 
(2) Starting from the origin point, the point in contour is determined by the distance and angle. 
(3) The angle is naturally directional and makes it very convenient to connect the points into a whole contour. We claim that Cartesian Representation may exhibit first two properties similarly. However, it lacks the advantage of the third property.      

We instantiate such an instance segmentation method by using the recent object detector FCOS~\cite{FCOS}, mainly for its simplicity.  Note that, it is possible to use other detectors such as RetinaNet~\cite{focalloss}, YOLO~\cite{yolo} with minimal modification to our framework. Specifically, we propose PolarMask, formulating instance segmentation as instance center classification and dense distance regression in a  polar coordinate, shown in Figure~\ref{fig:pipeline}. The model takes an input image and predicts the distance from a sampled positive  location (candidates of the instance center) to the instance contour at each angle, and after assembling, outputs the final mask. 
The overall pipeline of PolarMask is almost as simple and clean as FCOS. It introduces \textit{negligible} computation overhead. Simplicity and efficiency are the two key factors for single-shot instance segmentation, and PolarMask achieves them successfully.

Furthermore, PolarMask can be viewed as a generalization of FCOS. 
In other words, FCOS is a special case of PolarMask since bounding boxes can be viewed as the simplest mask with only 4 directions. Thus, one is suggested to use PolarMask over FCOS for instance recognition wherever mask annotation is available \cite{cityscapes,coco,pascalvoc,openimages}.  

In order to maximize the advantages of Polar Representation, we propose Polar Centerness and Polar IoU Loss to deal with sampling high-quality center examples and optimization for dense distance regression, respectively. They improve the mask accuracy by about  15\% relatively, showing considerable gains under stricter localization metrics. Without bells and whistles, PolarMask achieves 32.9\% in mask mAP with single-model and single-scale training/testing on the challenging COCO dataset~\cite{coco}.

The main contributions of this work are three-fold: 
\begin{itemize}
\itemsep -0.1cm
\item 
We introduce a brand new framework for instance segmentation, termed PolarMask, to model instance masks in the polar coordinate, which converts instance segmentation to two parallel tasks: 
instance center classification and dense distance regression. The main desirable characteristics of PolarMask is being simple and effective.

\item
We propose the Polar IoU Loss and Polar Centerness, tailored for our framework. 
We show that the proposed Polar IoU loss can largely ease the optimization and considerably improve the accuracy, compared with standard loss such as the smooth-\l1 loss. 
In parallel, Polar Centerness improves the original idea of ``Centreness'' in FCOS, leading to further performance boost.

\item 
For the first time, we show that the complexity of instance segmentation, in terms of both design and computation complexity, can be the same as bounding box object detection.
We further demonstrate this much simpler and flexible instance segmentation framework achieves competitive performance compared with more complex one-stage methods, which typically involve multi-scale training and longer training time. 
\end{itemize}

\section{Related Work}
\textbf{Two-Stage Instance Segmentation.}
Two-stage instance segmentation often formulates this task as the paradigm of ``Detect then Segment''~\cite{fcis,maskrcnn, panet,msrcnn}. They often detect bounding boxes then perform segmentation in the area of each bounding box. 
The main idea of FCIS \cite{fcis} is to predict a set of position-sensitive output channels fully convolutionally. These channels simultaneously address object classes, boxes, and masks, making the system fast.
Mask R-CNN~\cite{maskrcnn}, built upon Faster R-CNN, simply adds an additional mask branch and use RoI-Align to replace RoI-Pooling~\cite{fastrcnn} for improved accuracy.
Following Mask R-CNN, PANet~\cite{panet} introduces bottom-up path augmentation, adaptive feature pooling, and fully-connected fusion to boost up the performance of instance segmentation. 
Mask Scoring R-CNN~\cite{msrcnn} re-scores the confidence of mask from classification score by adding a mask-IoU branch, which makes the network to predict the IoU of mask and ground-truth.

In summary, the above methods typically consist of two steps, first detecting bounding box and then segmenting in each bounding box. They can achieve state-of-the-art performance but are often slow.

\vspace{2mm}
\textbf{One Stage Instance Segmentation.}
Deep Watershed Transform~\cite{deepwater} uses fully convolutional networks to predict the energy map of the whole image and use the watershed algorithm to yield connected components corresponding to object instances.
InstanceFCN~\cite{instancefcn} uses instance-sensitive score maps for generating proposals. It first produces a set of instance-sensitive score maps, then an assembling module is used to generate object instances in a sliding window.
The recent YOLACT~\cite{yolact} first generates  a set of prototype masks, the linear combination coefficients for each instance, and bounding boxes, then linearly combines the prototypes using the corresponding predicted coefficients and then crops with a predicted bounding box.
TensorMask~\cite{tensormask} investigates the paradigm of dense sliding-window instance segmentation, using structured 4D tensors to represent masks over a spatial domain.
ExtremeNet~\cite{extremenet} uses keypoint detection to predict 8 extreme points of one instance and generates an octagon mask, achieving relatively reasonable object mask prediction. 
The backbone of ExtremeNet is HourGlass~\cite{hourglass}, which is very heavy and often needs longer training time. 

Polar representation was 
firstly used 
in \cite{celldet} to detect cells in microscopic images, where the problem is much simpler as there are only two categories.
Concurrent to our work is 
the work of ESESeg~\cite{ese_seg}, which  also employs the polar coordinate to model instances.
However, our PolarMask achieves  significantly  better performance than ESESeg due to very different designs other than the polar representation.  
Note that most of these methods do not model instances directly and they can sometimes be hard to optimize~(e.g., longer training time, more data augmentation and extra labels). Our PolarMask directly models instance segmentation with a much simpler and flexible way of two paralleled branches: classifying each pixel of mass-center of instance and regressing the dense distance of rays between mass-center and contours. The most significant advantage of PolarMask is being simple and efficient compared with the above methods.

\section{Our Method}

In this section, we first briefly introduce the overall architecture of the proposed PolarMask. 
Then, we reformulate instance segmentation with the proposed Polar Representation. 
Next, we introduce a novel concept of Polar Centerness to ease the procedure of choosing high-quality center samples. 
Finally, we introduce a new Polar IoU Loss to optimize the dense regression problem.

\subsection{Architecture}

PolarMask is a simple, unified network composed of a backbone network~\cite{resnet}, a feature pyramid network~\cite{fpn}, and two or three task-specific heads,
depending on whether predicting bounding boxes.\footnote{It is optional to have the box prediction branch or not. As we empirically show,  the box prediction branch has little impact on mask prediction.} The settings of the backbone and feature pyramid network are the same as FCOS~\cite{FCOS}. 
While there exist many stronger candidates for those components, we align these settings with FCOS to show the simplicity and effectiveness of our instance modeling method.

\subsection{Polar Mask Segmentation}

In this section, we will describe how to model instances in the polar coordinate in detail.

\vspace{2mm}
\textbf{Polar Representation.} Given an instance mask, we firstly sample a candidate center $(x_{c}, y_{c})$ of the instance and the point located on the contour $(x_{i}, y_{i})$, $i = 1$, $2$, $...$, $N$. Then, starting from the center, $n$ rays are emitted uniformly with the same angle interval $\Delta\theta$ (e.g., $n=36$, $\Delta\theta=10^{\circ}$), whose length is determined from the center to the contour. In this way, we model the instance mask in the polar coordinate as one center and $n$ rays. Since the angle interval is pre-defined, only the length of the ray needs to be predicted. Therefore, we formulate the instance segmentation as instance center classification and dense distance regression in a polar coordinate.

\vspace{2mm}
\textbf{Mass Center.} There are many choices for the center of the instance, such as box center or mass-center. How to choose a better center depends on its effect on mask prediction performance. Here we verify the upper bound of box center and mass-center and conclude that mass-center is more advantageous. Details are in Figure~\ref{fig:upper bound}. We explain that the mass-center has a greater probability of falling inside the instance, compared with the box center. Although for some extreme cases, such as a donut, neither mass-center nor box center lies inside the instance. We leave it for further research.

\vspace{2mm}
\textbf{Center Samples.} Location $(x, y)$ is considered as a center sample if it falls into areas around the mass-center of any instance. Otherwise, it is a negative sample. We define the region for sampling positive pixels to be 1.5$\times$ strides~\cite{FCOS} of the feature map from the mass-center to left, top, right and bottom. 
Thus each instance has about 9$\sim $16 pixels near the mass-center as center examples. 
It has two advantages: 
(1) Increasing the number of positive samples from 1 to 9$\sim$16 can largely avoid imbalance of positive and negative samples. Nevertheless, focal loss~\cite{focalloss} is still needed
when training the classification branch. 
(2) Mass-center may not be the best center sample of an instance. More candidate points make it possible to automatically find the best center of one instance. We will discuss it in detail in Section~\ref{centerness}.

\begin{figure}[!t]
\begin{center}
\includegraphics[width=0.49\textwidth]{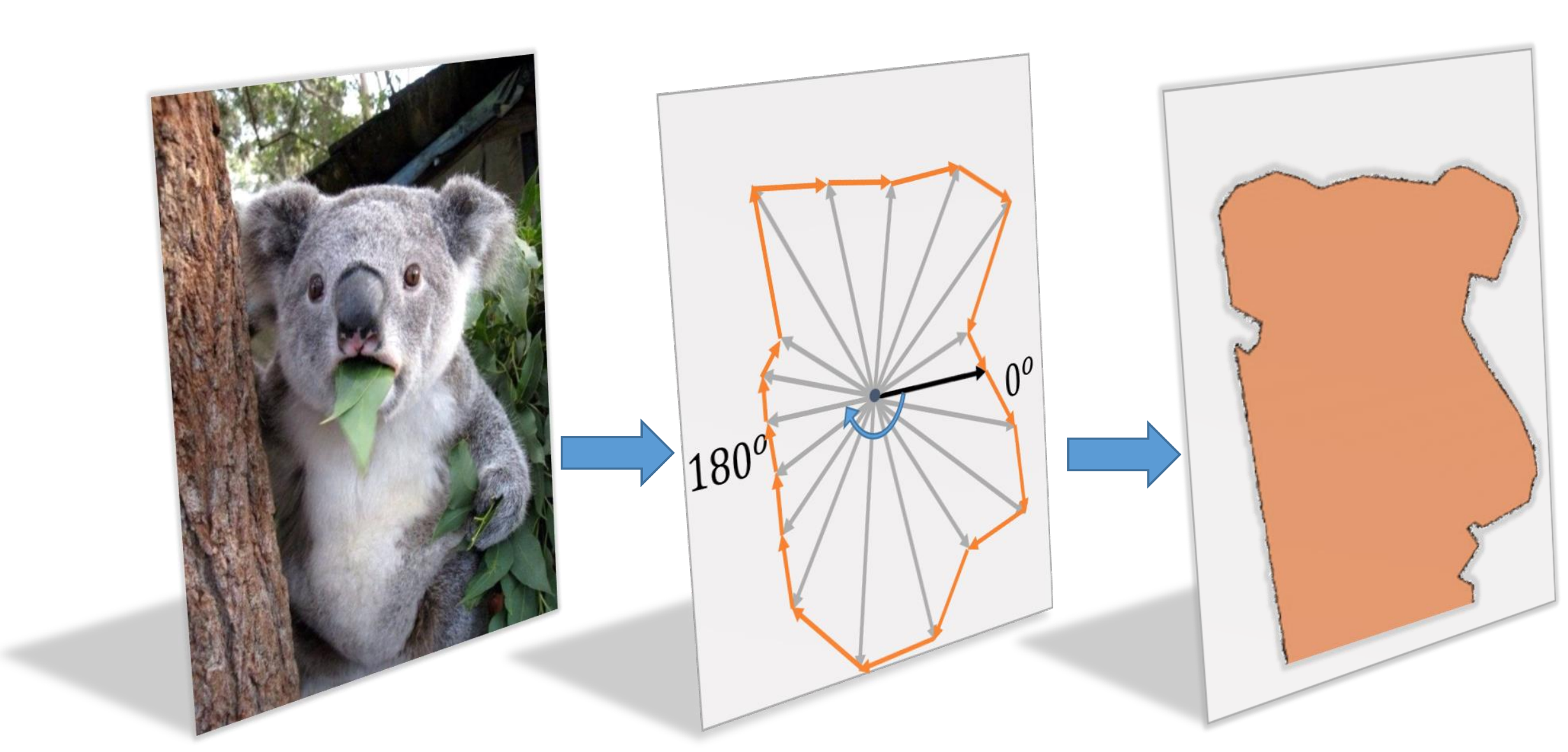}
\caption{\textbf{Mask Assembling}. Polar Representation provides a directional angle. The contour points are connected one by one start from $0^{\circ}$~(bold line) and assemble the whole contour and mask.}
\label{fig:polarseg}
\end{center}
\vspace{-5mm}
\end{figure}

\vspace{2mm}
\textbf{Distance Regression.} 
Given a center sample $(x_{c}, y_{c})$ and the contour point of an instance, the length of $n$ rays $\{d_{1}, d_{2}, \ldots, d_{n}\}$ can be computed easily. More details are in supplementary materials. Here we mainly discuss some corner cases:

\begin{itemize}
\itemsep -0.1cm

\item If one ray has multiple intersection points with the contour of instance, we directly choose the one with the maximum length.

\item If one ray, which starts from the center outside of the mask, does not have intersection points with the contour of an instance at some certain angles, we set its regression target as the minimum value $\epsilon$ (e.g., $\epsilon=10^{-6}$).

\end{itemize}

\begin{figure}[t]
\begin{center}
\includegraphics[width=0.49\textwidth]{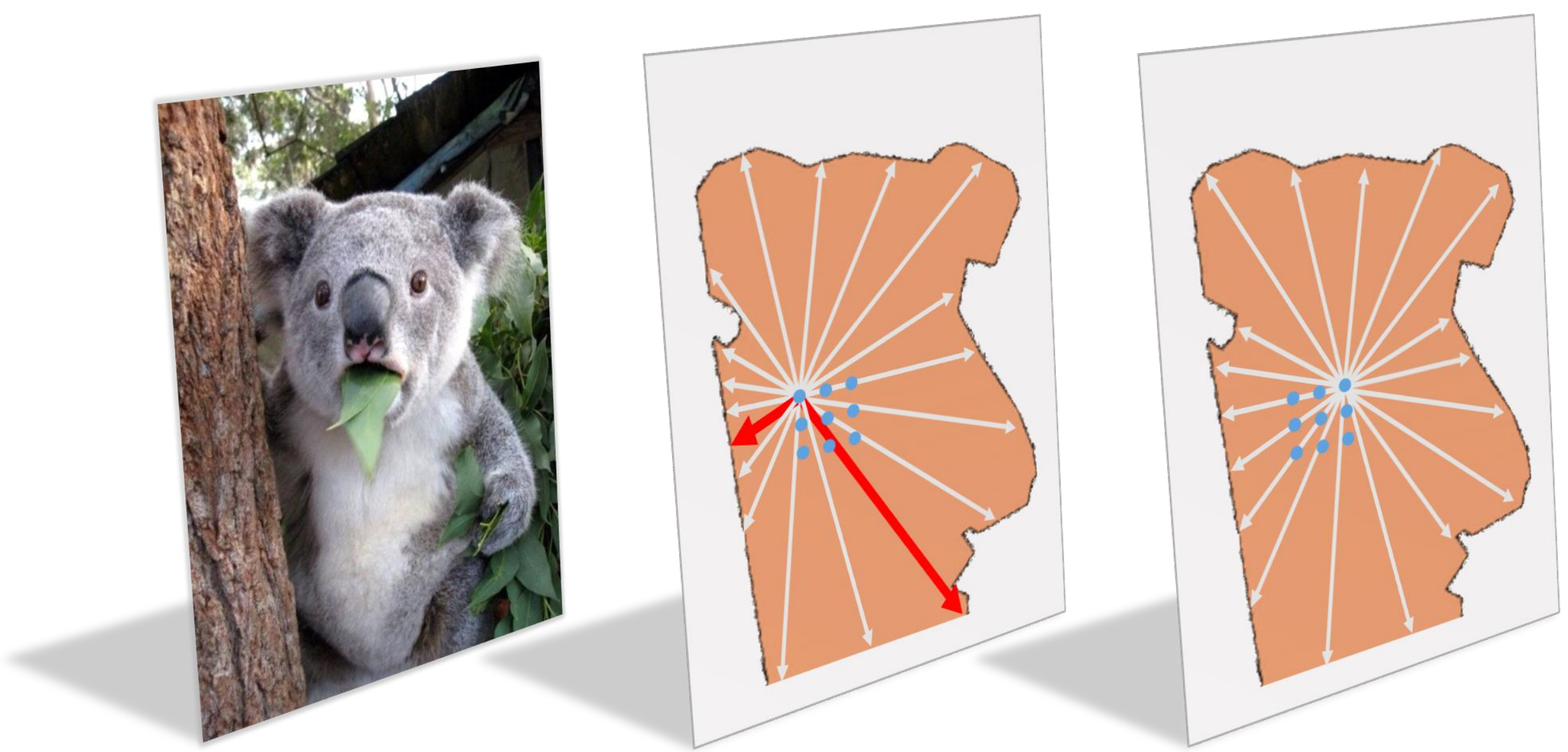}
\caption{\textbf{Polar Centerness}. Polar Centerness is used to down-weight such regression tasks as the high diversity of rays' lengths as shown in red lines in the middle plot. These examples are always hard to optimize and produce low-quality masks. During inference, the polar centerness predicted by the network is multiplied to the classification score, thus can down-weight the low-quality masks.}
\end{center}
\label{fig:centerness}
\vspace{-3mm}
\end{figure}

We argue that these corner cases are the main obstacles of restricting the upper bound of Polar Representation from reaching 100\% AP. However, it is not supposed to be seen as Polar Representation being inferior to the \textit{non-parametric} Pixel-wise Representation. The evidence is two-fold. First, even the Pixel-wise Representation still has certain gap with the upper bound of 100\% AP in practice, since some operation, such as down-sampling, is indispensable. Second, current  performance is far away from the upper bound regardless of the Pixel-wise Representation or Polar Representation. Therefore, the research effort is suggested to better spend on improving the practical performance of models, rather than  the theoretical upper bound. 

The training of the regression branch is non-trivial. First, the mask branch in PolarMask is actually a dense distance regression task since every training example has $n$ rays (e.g., $n=36$). It may cause an imbalance between the regression loss and classification loss. Second, for one instance, its $n$ rays are relevant and should be trained as a whole, rather than being seen as a set of independent regression examples. Therefore, we put forward the  Polar IoU Loss, discussed in detail in Section~\ref{iouloss}.

\vspace{2mm}
\textbf{Mask Assembling.} During inference, the network outputs the classification and centerness, we multiply centerness with classification  and obtain final confidence scores. We only assemble masks  from at most 1k top-scoring predictions per FPN level, after thresholding the confidence scores at 0.05. The top predictions from all levels are merged and non-maximum suppression (NMS) with a threshold of 0.5 is applied to yield the final results. Here we introduce the mask assembling process and a simple NMS process.

Given a center sample $(x_{c}, y_{c})$ and $n$ ray's length $\{d_{1}, d_{2}, \ldots, d_{n}\}$, we can calculate the position of each corresponding contour point with the following formula:

\begin{equation}
 x_{i} = \cos\theta_{i} \times d_i + x_{c}
\end{equation}
\begin{equation}
 y_{i} = \sin\theta_{i} \times d_i + y_{c}.
\end{equation}
Starting from $0^{\circ}$, the contour points are connected one by one, shown in Figure~\ref{fig:polarseg} and finally assembles a whole contour as well as the mask.

We apply NMS to remove redundant masks. To simplify the process, We calculate the smallest bounding boxes of masks and then apply NMS based on the IoU of generated boxes.

\subsection{Polar Centerness\label{centerness}}
Centerness~\cite{FCOS} is introduced to suppress these low-quality detected objects  without introducing any hyper-parameters and it is proven to be effective in object bounding box detection. However, directly transferring it to our system can be sub-optimal since its centerness is designed for bounding boxes and we care about mask prediction. 

Given a set $\{d_{1}, d_{2}, \ldots, d_{n}\}$ for the length of $n$ rays of one instance.
We propose Polar Centerness: 
\begin{equation}
{ \rm Polar~Centerness } 
= \sqrt{\frac{\min(\{d_{1}, d_{2}, \ldots, d_{n}\})}
        {\max(\{d_{1}, d_{2}, \ldots, d_{n}\})}}
\end{equation}
It is a simple yet effective strategy to re-weight the points so that the closer $d_{min}$ and $d_{max}$ are, higher weight the point is assigned.

We add a single layer branch, in parallel with the classification branch to predict Polar Centerness of a location, as shown in Figure~\ref{fig:pipeline}. Polar Centerness predicted by the network is multiplied to the classification score, thus can down-weight the low-quality masks.  Experiments show that Polar Centerness improves accuracy especially under stricter localization metrics, such as AP$_{75}$.

\subsection{Polar IoU Loss \label{iouloss}}
As discussed above, the method of polar segmentation  converts the task of instance segmentation into a set of regression problems. In most cases in the field of object detection and segmentation, smooth-\l1 loss~\cite{rcnn} and IoU loss~\cite{unitbox} are the two effective ways to supervise the regression problems. Smooth-\l1 loss overlooks the correlation between samples of the same objects, thus, resulting in less accurate localization. IoU loss, however, considers the optimization as a whole, and directly optimizes the metric of interest, IoU. Nevertheless, computing the IoU of the predicted mask and its ground-truth is tricky and very difficult to implement parallel computations. In this work, we derive an easy and effective algorithm to compute mask IoU based on the polar vector representation and achieve competitive performance.

\begin{figure}[bt!]
\centering 
\includegraphics[width=0.49\textwidth]{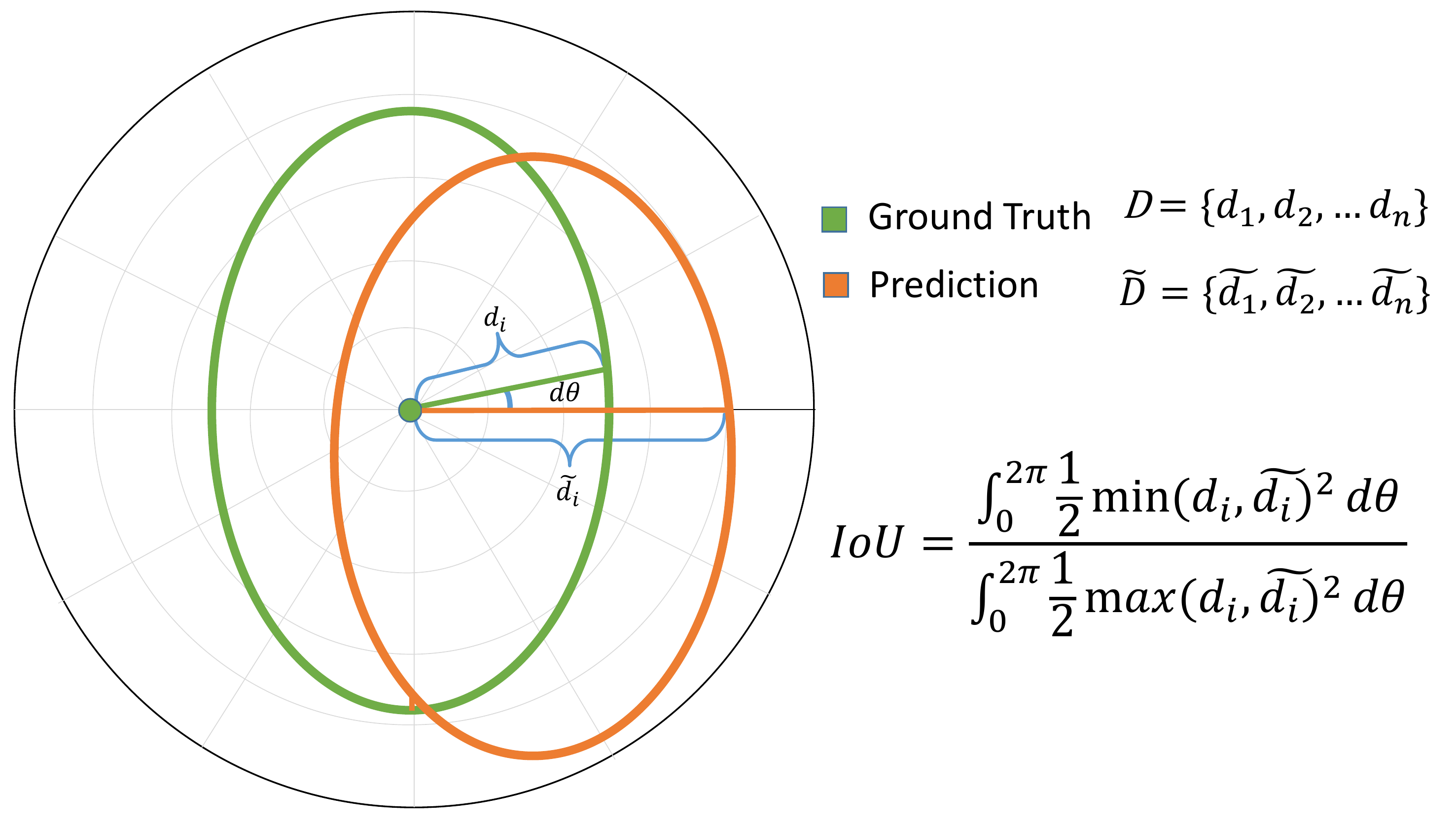}
\caption{\textbf{Mask IoU in Polar Representation}. Mask IoU (interaction area over union area) in the polar coordinate can be calculated by integrating the differential IoU area in terms of differential angles.}
\label{fig_iouloss}
\vspace{-5mm}
\end{figure}

We introduce Polar IoU Loss starting from the definition of IoU, which is the ratio of interaction area over union area between the predicted mask and ground-truth. As shown in Figure~\ref{fig_iouloss}, in the polar coordinate system, for one instance, mask IoU is calculated as follows:
\begin{equation}
  { \rm IoU  } = \frac{\int_{0}^{2\pi} \frac{1}{2} \min(d,d^*)^2 d\theta}
        {\int_{0}^{2\pi}\frac{1}{2} \max(d,d^*)^2 d\theta}
\end{equation}
where regression target $d$ and predicted $d^*$ are length of the ray, angle is $\theta$.
Then we transform it to the discrete form\footnote{For notation convenience, we define:
\begin{equation}
    d_{\min}=\min(d,d^*), d_{\max}=\max(d,d^*).
\end{equation}
}
\begin{equation}
 {  \rm IoU } = \lim_{N \to \infty}\frac
  {\sum_{i=1}^{N}\frac{1}{2} d_{\min}^2 \Delta \theta_i}
  {\sum_{i=1}^{N}\frac{1}{2} d_{\max}^2 \Delta \theta_i}
\end{equation}
When $N$ approaches infinity, the discrete form is equal to continuous form. We assume that the rays are uniformly emitted, so $\Delta \theta = \frac{2\pi}{N}$, which further simplifies the expression. We empirically observe that the power form has little impact on the performance~($\pm0.1$ mAP difference) if it is discarded and simplified into the following form:
\begin{equation}
  {\rm Polar~IoU}  = \frac{\sum_{i=1}^{n}d_{\min}}
        {\sum_{i=1}^{n}d_{\max}}
\end{equation}

Polar IoU Loss is the binary cross entropy (BCE) loss of Polar IoU. Since the optimal IoU is always 1, the loss is actually is negative logarithm of Polar IoU:
\begin{equation}
{  \rm Polar~IoU~Loss }
= \log \frac{\sum_{i=1}^{n}d_{\max}}
        {\sum_{i=1}^{n}d_{\min}}
\end{equation}

\begin{table*}[t]	
	\centering
    \input{tables/table1.tex}

	\caption{Ablation experiments for PolarMask. All models are trained on $\tt trainval35k$ and tested on $\tt  minival$, using ResNet50-FPN backbone unless otherwise noted.}
	\label{table:1}
	\vspace{-3mm}
\end{table*}

Our proposed Polar IoU Loss exhibits two advantageous properties: (1) It is differentiable, enabling back propagation; and it is very easy to implement parallel computations, thus facilitating a fast training process. (2) It predicts the regression targets as a whole. It improves the overall performance by a large margin compared with smooth-\l1 loss, shown in our experiments. (3) As a bonus, Polar IoU Loss is able to automatically keep the balance between classification loss and regression loss of dense distance prediction. We will discuss it in detail
in our experiments.

\section{Experiments}
We present results of instance segmentation on the challenging COCO benchmark~\cite{coco}. Following common practice~\cite{maskrcnn,tensormask}, we train using the union of 80K train images and a 35K subset of val images ($\tt trainval35k$), and report ablations on the remaining 5K val.\  images ($\tt minival$). We also compare results on $\tt test$-$\tt dev$. We adopt the 1$\times$ training strategy~\cite{Detectron,mmdetection}, single scale training and testing of image short-edge as 800 unless otherwise noted. 

\textbf{Training Details.}
In ablation study, ResNet-50-FPN~\cite{resnet,fpn} is used as our backbone networks and the same hyper-parameters with FCOS~\cite{FCOS} are used. Specifically, our network is trained with stochastic gradient descent (SGD) for 90K iterations with the initial learning rate being 0.01 and a mini-batch of 16 images. The learning rate is reduced by a factor of 10 at iteration 60K and 80K, respectively. Weight decay and momentum are set as 0.0001 and 0.9, respectively.
We initialize our backbone networks with the weights pre-trained on ImageNet~\cite{imagenet}. The input images are resized to have their shorter side being 800 and their longer side less or equal to 1333.

\subsection{Ablation Study}
\label{exp}
\begin{figure}[t]
\begin{center}
\includegraphics[width=0.49\textwidth]{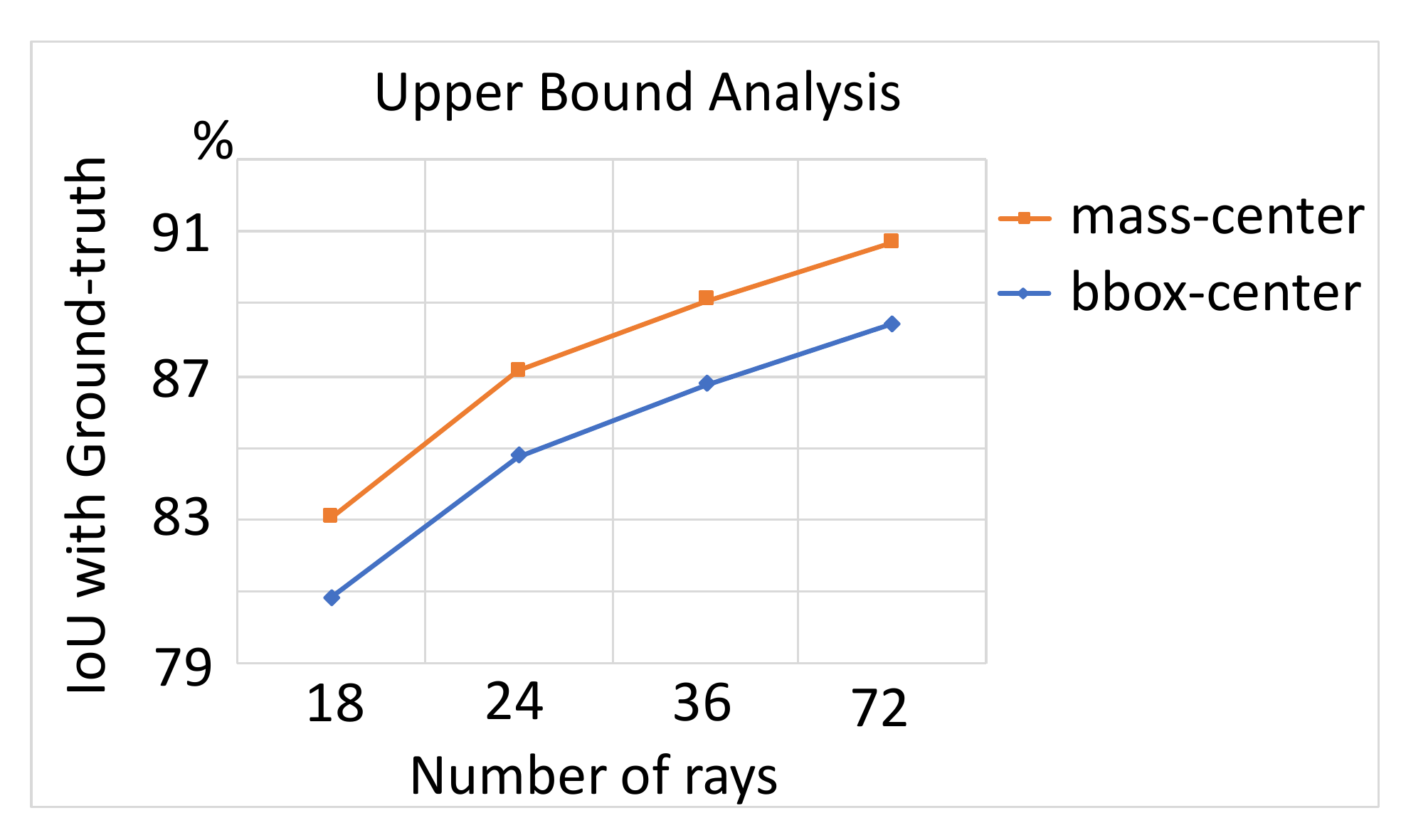}
\caption{Upper Bound Analysis. More rays can model instance mask with higher IoU with Ground Truth, and mass-center is more friendly to represent an instance than box-center. With
more rays~, \eg 90 rays improve 
0.4\% compared to 72 rays; and result is saturated with 120 rays.}
\label{fig:upper bound}
\end{center}
\vspace{-5mm}
\end{figure}

\textbf{Verification of Upper Bound.} The first concern about PolarMask is that it might not depict the mask precisely. In this section, we prove that this concern may not be necessary. Here we verify the upper bound of PolarMask as the IoU of predicted mask and ground-truth when all of the rays regress to the distance equal to ground-truth. The verification results on different numbers of rays are shown in Figure~\ref{fig:upper bound}. It can be seen that IoU is approaching to nearly perfect~(above 90\%) when the number of rays increases, which shows that Polar Segmentation is able to model the mask very well. Therefore, the concern about the upper bound of PolarMask is not necessary. Also, it is more reasonable to use mass-center than bounding box-center as the center of an instance because the bounding box center is more likely to fall out of the instance.

\begin{figure*}[t]
\begin{center}
\includegraphics[width=0.97\textwidth]{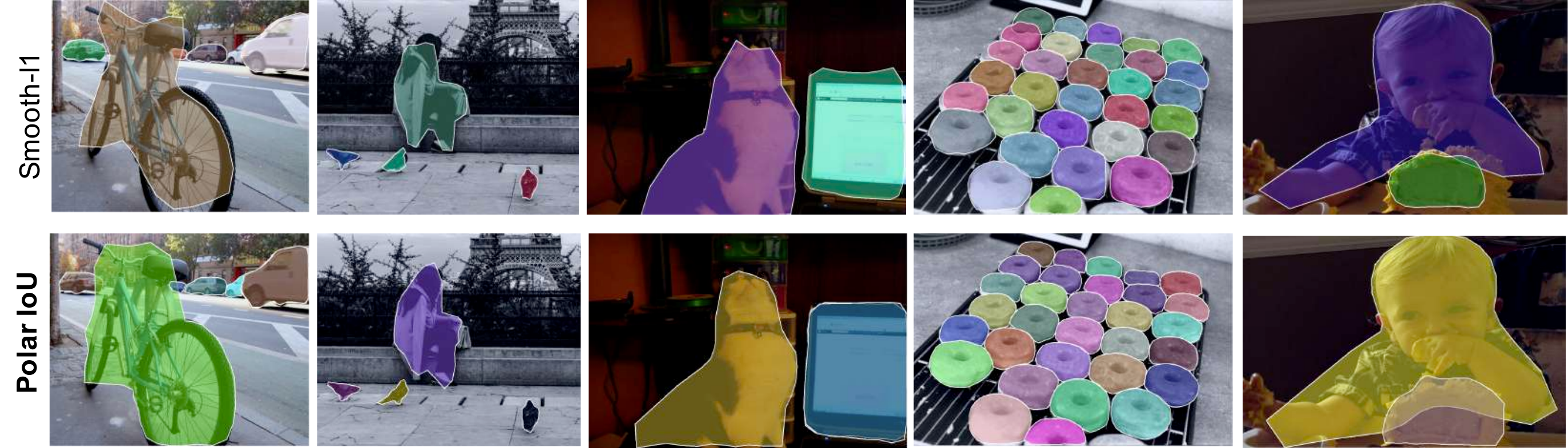}
\caption{Visualization of PolarMask with Smooth-\L1 loss and Polar IoU loss. Polar IoU Loss achieves to regress more accurate contour of instance while Smooth-\L1 Loss exhibits systematic artifacts.}
\label{fig:giraffe}
\end{center}
\vspace{-5mm}
\end{figure*}

\textbf{Number of Rays.} It plays a fundamental role in the whole system of PolarMask. From Table~\ref{table:vectors} and Figure~\ref{fig:upper bound}, more rays show higher upper bound and better AP. For example, 36 rays improve by 1.5\% AP compared to 18 rays. Also, too many rays, 72 rays, saturate the performance since it already depicts the mask contours well and the number of rays is no longer the main factor constraining the performance. 

\textbf{Polar IoU Loss \emph{vs}. Smooth-\L1 Loss.} We test both Polar IoU Loss and Smooth-\L1 Loss in our architecture. We note that the regression loss of Smooth-\L1 Loss is \textit{significantly} larger than the classification loss since our architecture is a task of dense distance prediction. To cope with the imbalance, we select different factor $\alpha$ to regression loss in Smooth-\L1 Loss. Experiment results are shown in Table~\ref{table:loss}. Our Polar IoU Loss achieves 27.7\%  AP without balancing regression loss and classification loss. In contrast, the best setting for Smooth-\L1 Loss achieves 25.1\%  AP, a gap of 2.6\% AP, showing that Polar IoU Loss is more effective than Smooth-\L1 loss for training the regression task of distances between mass-center and contours. 

We hypothesize that the gap may come from two folds. First, the Smooth-\L1 Loss may need more hyper-parameter search to achieve better performance, which can be time-consuming compared to the Polar IoU Loss. Second, Polar IoU Loss predicts all rays of one instance as a whole, which is superior to Smooth-\L1 Loss.  

In Figure~\ref{fig:giraffe} we compare some results using the  Smooth-\L1 Loss and Polar IoU Loss respectively. Smooth-\L1 Loss exhibits systematic artifacts, suggesting that it lacks supervision of the level of the whole object. PolarMask shows more smooth and precise contours.

\textbf{Polar Centerness \emph{vs}. Centerness.} Visualization results can be found in the supplementary material. The comparison experiments are shown in Table~\ref{table:centerness}. Polar Centerness improves by 1.4\%  AP overall. 

Particularly, AP$_{75} $ and AP$_{L}$ are raised considerably, 2.3\%  AP and 2.6\%  AP, respectively. We explain as follows. On the one hand, low-quality masks make more negative effect on high-IoU. On the other hand, large instances have more possibility of large difference between maximum and minimum lengths of rays, which is exactly the problem that Polar Centerness is committed to solve.

\begin{figure*}[h!]
\begin{center}
\includegraphics[width=0.95\textwidth]{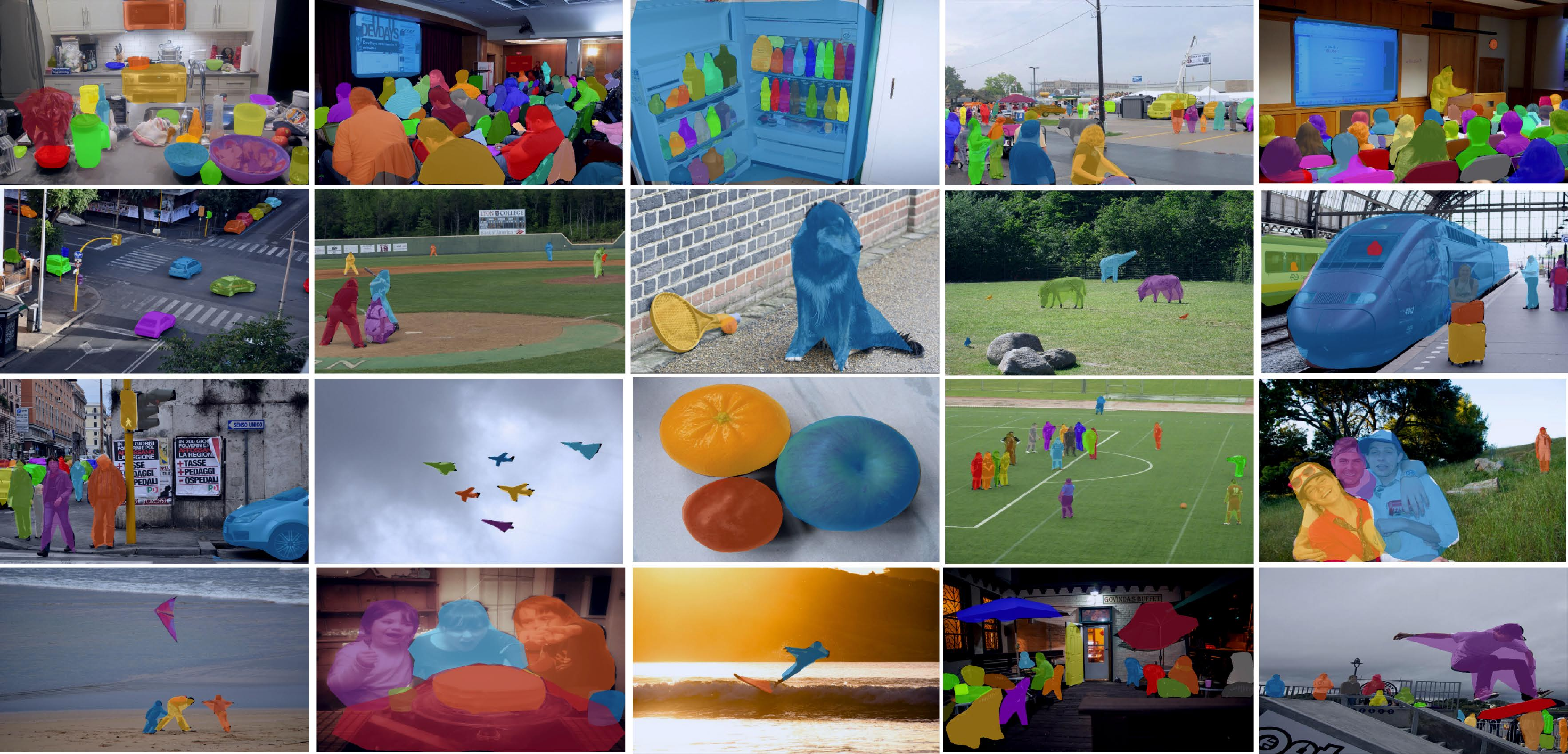}
\vspace{2mm}
\caption{Results of PolarMask on COCO $\tt test$-$\tt dev$ images with ResNet-101-FPN,  achieving 30.4\%  mask AP (Table~\ref{table:2}).}
\label{fig:result}
\end{center}
\vspace{2mm}
\end{figure*}

\begin{table*}[t]
    \centering
    \input{tables/tables2.tex}
    \vspace{2mm}
	\caption{\textbf{Instance segmentation} mask AP on the COCO $\tt test$-$\tt dev$. The standard training strategy~\cite{Detectron} is training by 12 epochs; and `aug' means data augmentation, including multi-scale and random crop. $\checkmark$ is training with `aug', $\circ$ is without `aug'. }
	\label{table:2}
\end{table*}

\textbf{Box Branch.} Most of previous methods of instance segmentation require the bounding box to locate area of object and then segment the pixels inside the object. In contrast, \textit{PolarMask is capable to directly output the mask without bounding box}. 

In this section, we test whether the additional bounding box branch can help improve the mask AP as follows. 
From Table~\ref{table:box}, we can see that bounding box branch makes little difference to performance of mask prediction. Thus, we do not have the bounding box prediction head in PolarMask for simplicity and faster speed.

\textbf{Backbone Architecture}. Table~\ref{table:backbone} shows the results of PolarMask on different backbones. It can be seen that better feature extracted by deeper and advanced design networks improve the performance as expected.

\textbf{Speed \emph{vs}. Accuracy.} Larger image sizes yield higher accuracy, in slower inference speeds. Table~\ref{table:speed} shows the speed and accuracy trade-off for different input image scales, defined by the shorter image side. 
The FPS is reported on one V100 GPU. Note that here we report the entire inference time, all post-processing included. 
It shows that PolarMask has a strong potentiality to be developed as a real-time instance segmentation application with simple modification. 
We also report more results in different benchmarks in the supplementary material.

\subsection{Comparison against state-of-the-art}
We evaluate PolarMask on the COCO dataset and compare $\tt test$-$\tt dev$ results to state-of-the-art methods including both one-stage and two-stage models, shown in Table~\ref{table:2}. PolarMask outputs are visualized in Figure~\ref{fig:result}.
For data augmentation, we randomly scale the shorter side of images in the range from 640 to 800 during the training. 

Without any bells and whistles, PolarMask is able to achieve competitive performance with more complex one-stage methods. With a simpler pipeline and half training epochs, PolarMask outperforms YOLACT with 0.9 mAP. Moreover, the best PolarMask with deformable convolutional layers~\cite{dcn} can achieve 36.2 mAP, which is comparable with state-of-the-art methods.

In the supplementary material, we compare the FPS between TensorMask and PolarMask with the same image size and device. PolarMask can run at 12.3 FPS with the ResNet-101 backbone, which is 4.7 times faster than TensorMask. Even 
when 
equipped with DCN~\cite{dcn}, PolarMask can still be three times faster than TensorMask.

\section{Conclusion}
PolarMask is a single shot anchor-box free instance segmentation method. Different from previous works that typically solve mask prediction as binary classification in a spatial layout, PolarMask puts forward to represent a mask by its contour and model the contour by one center and rays emitted from the center to the contour in polar coordinate. PolarMask is designed almost as simple and clean as single-shot object detectors, introducing negligible computing overhead. We hope that the proposed PolarMask framework can serve as a fundamental and strong baseline for single-shot instance segmentation tasks.

\appendix 
\section{Appendix}
\subsection{Distance Label Generation}
Here we explain 
the detail of distance label generation in Algorithm~\ref{alg:distance}. Firstly, we get the contours of one instance. Methods like $\tt cv2.findContours$ in OpenCV can be applied to obtain the contour. Second we traverse every point on the contour to calculate the distance and the angle from this point to the center of instance. Thirdly we take out distance with corresponding angle 
(e.g., 36 rays, $\Delta\theta=10^{\circ}$). 
Note that besides two corner cases discussed in the original paper, there remains another situation:
The intersection point between a ray and the contour happens to be a sub-pixel (i.e., pixel coordinates are not integers), in this case the target angle misses and we can find the nearest angle to replace it. For instance, if $10^{\circ}$ and corresponding distance miss, however, $9^{\circ}$ exists, we can use $9^{\circ}$ as its regression target.

\begin{algorithm}[h]
		\footnotesize 
		\caption{Distance Label Generation~(e.g., 36 rays)}
		\begin{algorithmic}[1]
			\Require Contour: $Contour$, Center Sample: $center$, 
			\Function {Distance Calculate}{$Contour$, $center$}
			\State Initialize distance set D, angle set A
			\For{each $point \in Contour$}
			    \State Calculate distance and angle from $point$ to $center$
			    \State Append distance to D, angle to A
			\EndFor
			\State Get distance set D, angle set A  
			\\
			\State Initialize distance label L$_D$
            \For{angle $\theta$ $\in$ [0,10,20,\dots,360]}
                \If {Find angle $\theta$ in A}
                    \If {angle has multiple distances $d$}  
                        \State Find the maximum $d$
                    \Else
			            \State Find corresponding distance $d$
			        \EndIf
    			\Else {~~$\theta$ not in A}
    			    \If {Find angle $\theta_{near}$ nearby $\theta$ in A}
    			        \State Find corresponding $d$    \ \ \ \ \  // Nearest Interpolation.
    			    \Else
    			        \State $d$ = $10^{-6}$        \ \ \ \ \ \ \ \ \ \ \ \ \ // Target a minimum number as label.
    			    \EndIf
    			\EndIf
    			\State Append $d$ to L$_D$
    		\EndFor
    		\State \Return{L$_D$}
			\EndFunction
			
		\end{algorithmic}
		\label{alg:distance}
	\end{algorithm}

\subsection{Computation Complexity and Speed Analysis}
In this section, we compare the complexity and speed of different methods. All the settings are
the same for fair comparison for all methods.

Table~\ref{table:5} shows the computation complexity and parameters of different methods. We set the input image size equals $800*1280$ and ResNet50 as the backbone for all methods. Note that PolarMask without box branch introduces marginal cost when compared with object detector FCOS in both computation complexity and parameters.

Table~\ref{table:6} compare the speed of different methods. The testing time includes the model inference and post-processing, such as Non-Maximum Suppression~(NMS). The test images are resized to have their shorter side being 800 and their longer side less or equal to 1333.
PolarMask is faster than the one-stage method TensorMask and two-stage method Mask R-CNN due to the simple pipeline design. Even 
when
equipped with DCN PolarMask can  still be three times faster than TensorMask.

\begin{table}[h]
    \centering
    \scalebox{0.77}{
    \input{tables/table4.tex}}
	\caption{Computation complexity and parameters comparison with other methods. Note that ``PolarMask w/o box'' only introduces marginal computation when compared with FCOS.}
	\label{table:5}
\vspace{-3mm}
\end{table}

\begin{table}[h]
    \centering
    \scalebox{0.9}{
    \input{tables/table5.tex}}
	\caption{Speed analysis of different methods. All post-processing are included. 
	The input images are resized to have their shorter side being 800 and their longer side less or equal to 1333.
	PolarMask has large advantages than TensorMask in speed even equipped with deformable convolutional operations.}
	\label{table:6}
\vspace{-3mm}
\end{table}

\begin{figure*}[t]
\begin{center}
\includegraphics[width=0.99\textwidth]{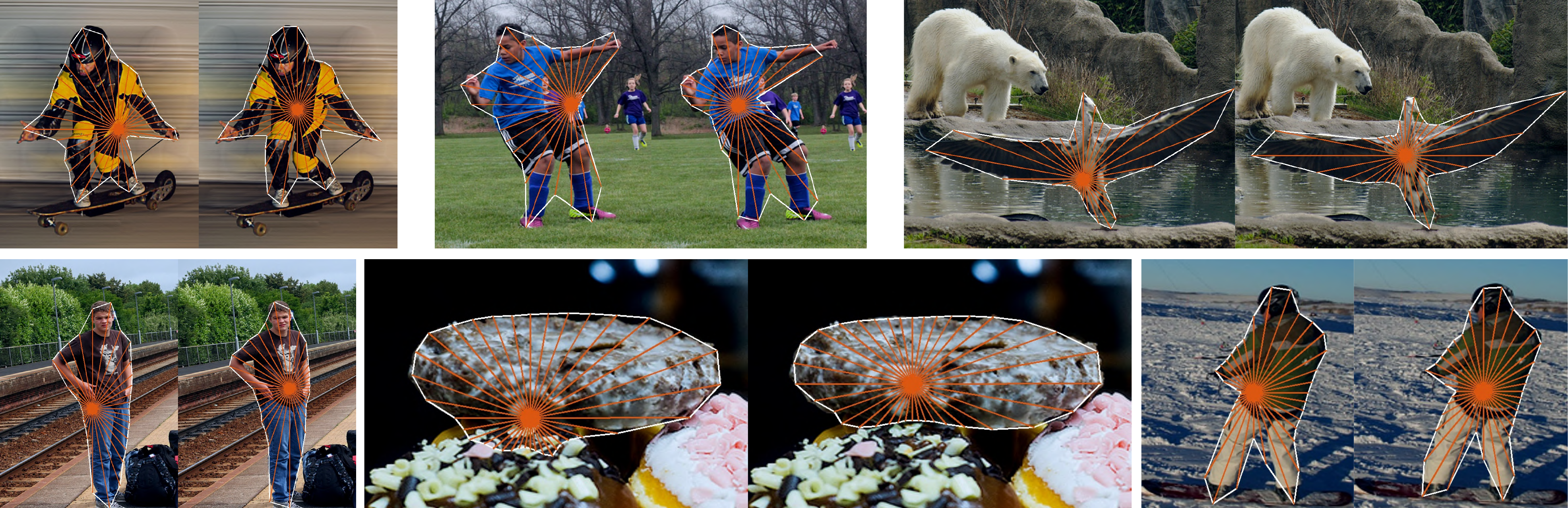}
\caption{Visualization of Cartesian Centerness and Polar Centerness. Left:~Cartesian Centerness; Right:~Polar Centerness. Orange lines are predicted distance of rays, emitting from the center to the contour. White lines are the contours of objects.}
\label{fig:polar_centerness}
\end{center}
\vspace{-3mm}
\end{figure*}

\subsection{Polar Centerness \emph{vs}. Centerness}
We visualize the predicted results of PolarMask trained with Polar Centerness and original Centerness in Fig.~\ref{fig:polar_centerness}. Visualized images demonstrate that Polar Centerness can optimize the model to automatically increase the weight of high-quality positive samples while decreasing the weight of low-quality positive samples in instance segmentation tasks, and the distances regression of rays are more accuracy. Thanks to Polar Centerness, the high-quality center will make the network more easily to regress because of the distances of different rays are more balanced under the constraint of Polar Centerness. 

\subsection{More Benchmark Results for PolarMask}
We present more benchmark results of our proposed PolarMask in Table~\ref{table:4}. All models were tested on the MS-COCO \cite{coco} validation set~($\tt minival$). ``DCN'' denotes deformable convolution layers~\cite{dcn} in backbone and head. ``ms-train'' means we add multi-scale training strategy, same as TensorMask~\cite{tensormask}, randomly scaling the short side of the image from 640 to 800.

Multi-scale training can improve the final results at 1\%-1.5\%. And DCN can boost up to at least 2.3\% at the different backbones. Note that the best PolarMask with ResNext-101, DCN, and multi-scale training achieves 35.9 mAP.

\subsection{Comparison with ESE-Seg}
PolarMask and ESE-Seg are concurrent works and conducted independently. 
Although ESE-Seg also used polar representation, we propose `Polar IoU Loss' and `Polar Centerness', which are two modules that were not presented in ESE-Seg. They're unique in PolarMask and crucial to improve performance (ours outperforms ESE-Seg by 7.5 AP).

First, 
ESE-Seg requires box detection, which is not necessary in PolarMask. This is a significant difference. 
Second, ESE-Seg sets the inner center as positive sample. But PolarMask applies `Polar Centerness' to automatically choose the best sample among multiple positive samples near mass center.
Third, ESE-Seg regresses coefficients after Chebyshev polynomial fitting and trains with traditional L2 loss, while PolarMask regresses lengths directly using the proposed `Polar IoU Loss'.

As a result, PolarMask significantly outperforms ESE-Seg.
Table~\ref{table1} compares PolarMask with ESE-Seg on COCO 2017 val. It demonstrates PolarMask improves performance significantly and surpasses ESE-Seg by nearly 7.5 AP.

\begin{table}[t]
    \centering
    \input{tables/table3.tex}
	\caption{Benchmark results of PolarMask on MS-COCO~\cite{coco} validation set ($\tt minival$). All the models here are trained with FPN~\cite{fpn}. For the backbone notation, ``R-50'' and ``R-101'' denotes ResNet-50 and ResNet-101~\cite{resnet} respectively. ``DCN'' denotes  deformable convolution layers~\cite{dcn} in backbone and head. ``X'' denotes the ResNeXt-101~\cite{resnext} backbone. ``ms'' indicates multi-scale. }
	\label{table:4}
\end{table}

\begin{table}[h]
    \centering
    \input{tables/table6.tex}
	\caption{Comparasion of PolarMask and ESE-Seg on COCO 2017 val. $\checkmark$ is equipped with `Polar IoU Loss' and `Polar Centerness', $\circ$ is not.}
	\label{table1}
\end{table}

\textbf{Acknowledgements and Declaration of Conflicting Interests}
Chunhua Shen and his employer received no financial support for the research, authorship, and/or publication of this article.

{\small
\bibliographystyle{ieee_fullname}
\bibliography{polarmask}
}

\end{document}

%% file: tables/table1.tex
% number of rays
\begin{subtable}[t]{3.2in}
	\centering
	\setlength{\tabcolsep}{1.5mm}
	\small 
	\begin{tabular}{c|ccc|ccc}
	rays & AP & AP$_{50}$ & AP$_{75}$ & AP$_{S}$ & AP$_{M}$ & AP$_{L}$ \\
	\hline
	18 & 26.2 & 48.7 & 25.4 &11.8 &28.2 &38.0\\
	24 & 27.3 & 49.5 & 26.9 &12.4 &29.5 &40.1\\
	\textbf{36} & \textbf{27.7} &49.6  &  27.4 &  12.6 & 30.2 & 39.7\\
	72 & 27.6 &  49.7 & 27.2 & 12.9 &30.0 & 39.7\\
	\end{tabular}
	\caption{\textbf{Number of Rays}: More rays bring a large gain, while too many rays saturate since it already depicts the mask ground-truth well.}
	\label{table:vectors}
	\vspace{5mm}
\end{subtable}
\quad
% loss
\begin{subtable}[t]{3.2in}
    \centering
	\setlength{\tabcolsep}{1.0mm}
	\small 
	\begin{tabular}{c|c|ccc|ccc}
	loss & $\alpha$  & AP & AP$_{50}$ & AP$_{75}$ & AP$_{S}$ & AP$_{M}$ & AP$_{L}$\\
	\hline
	\multirow{3}*{Smooth-\L1} &
	0.05 & 24.7 & 47.1 & 23.7 &11.3 &26.7 &36.8\\
	& 0.30 & 25.1 & 46.4 & 24.5 &10.6 &27.3 &37.3\\
	& 1.00 & 20.2 & 37.9 & 19.6 &8.6 &20.6 &31.1\\
	\hline
	\textbf{Polar IoU} & 1.00  & \textbf{27.7} & 49.6 & 27.4 &12.6 &30.2 &39.7 \\
	\end{tabular}
	\caption{\textbf{Polar IoU Loss \emph{vs}. Smooth-L1 Loss}: Polar IoU Loss outperforms Smooth-\L1 loss, even the best variants of balancing regression loss and classification loss.}
	\label{table:loss}
	\vspace{5mm}
\end{subtable}
\quad
% centerness
\begin{subtable}[t]{3.2in}
	\centering
	\setlength{\tabcolsep}{1.2mm}
	\small 
	\begin{tabular}{c|ccc|ccc}
	centerness & AP & AP$_{50}$ & AP$_{75}$ & AP$_{S}$ & AP$_{M}$ & AP$_{L}$\\
	\hline
	Original & 27.7 & 49.6 & 27.4 & 12.6&30.2& 39.7\\
	\textbf{Polar} & \textbf{29.1} & 49.5 & 29.7 & 12.6 & 31.8 & 42.3\\
	\end{tabular}
	\caption{\textbf{Polar Centerness \emph{vs}.  Centerness}: Polar Centerness bring a large gain, especially high IoU AP$_{75}$ and large instance AP$_L$.}
	\label{table:centerness}
	\vspace{5mm}
\end{subtable}
\quad
% box branch
\begin{subtable}[t]{3.2in}
	\centering
	\setlength{\tabcolsep}{1.2mm}
	\small 
	\begin{tabular}{c|ccc|ccc}
	box branch & AP & AP$_{50}$ & AP$_{75}$ & AP$_{S}$ & AP$_{M}$ & AP$_{L}$\\
	\hline
	w & 27.7 & 49.6 & 27.4 & 12.6&30.2&39.7\\
	w/o & 27.5 & 49.8 & 27.0 &13.0 &30.0 &40.0\\
	\end{tabular}
	\caption{\textbf{Box Branch}: Box branch makes no difference to performance of mask prediction.}
	\label{table:box}
	\vspace{5mm}
\end{subtable}
\quad
% backbone
\begin{subtable}[t]{3.2in}
	\centering
	\setlength{\tabcolsep}{1.5mm}
	\small 
	\begin{tabular}{l|ccc|ccc}
	backbone & AP & AP$_{50}$ & AP$_{75}$ & AP$_{S}$ & AP$_{M}$ & AP$_{L}$\\
	\hline
	ResNet-50 & 29.1 & 49.5 & 29.7  & 12.6 & 31.8 & 42.3 \\
	ResNet-101 & 30.4 & 51.1 & 31.2 & 13.5 & 33.5 & 43.9 \\
	ResNeXt-101 & 32.6 & 54.4 & 33.7 & 15.0 & 36.0 & 47.1 \\
	\end{tabular}
	\caption{\textbf{Backbone Architecture}: All models are based on FPN. Better backbones bring expected gains: deeper networks do better, and ResNeXt improves on ResNet.}
	\label{table:backbone}
\end{subtable}
\quad
% scale
\begin{subtable}[t]{3.2in}
	\centering
	\setlength{\tabcolsep}{1.0mm}
	\small 
	\begin{tabular}{l|ccc|ccc|c}
	scale & AP & AP$_{50}$ & AP$_{75}$ & AP$_{S}$ & AP$_{M}$ & AP$_{L}$ &  FPS\\
	\hline
	400 & 22.9 & 39.8 & 23.2  & 4.5 & 24.4 & 41.7 &  26.3\\
	600 & 27.6 & 47.5 & 28.3 & 9.8 & 30.1 & 43.1 &   21.7\\
	800 & 29.1 & 49.5 & 29.7  & 12.6 & 31.8 & 42.3 & 17.2\\
	\end{tabular}
	\caption{\textbf{Accuracy/speed trade-off on ResNet-50}: PolarMask performance with different image scales. 
% 	The FPS is reported on outdated Maxwell TitanX GPUs and V100.
    The FPS is reported on one V100 GPU.
	%For Titan-X, the speed is about 50\% slower than 1080Ti.
	}
	\label{table:speed}
\end{subtable}

%% file: tables/tables2.tex
\def\x{{$\footnotesize \times$}}
% \small 
\begin{tabular}{l|l|cc|ccc|ccc}
	method & backbone & epochs & aug & AP & AP$_{50}$ & AP$_{75}$ & AP$_{S}$ & AP$_{M}$ & AP$_{L}$\\
	\hline
    \emph{two-stage} &&&&&&&&\\
    MNC~\cite{mnc} & ResNet-101-C4 & 12 & $\circ$ & 24.6 & 44.3 & 24.8 & 4.7 & 25.9 & 43.6\\
	FCIS~\cite{fcis} & ResNet-101-C5-dilated & 12 & $\circ$ & 29.2 & 49.5 & - & 7.1 & 31.3 & 50.0\\
	Mask R-CNN~\cite{maskrcnn} & ResNeXt-101-FPN & 12 &$\circ$ & 37.1 & 60.0 & 39.4 & 16.9 & 39.9 & 53.5\\
	\hline
	\emph{one-stage} &&&&&&&&&\\
	ExtremeNet~\cite{extremenet}  & Hourglass-104 & 100 &\checkmark &  18.9  & 44.5 & 13.7 & 10.4 & 20.4 & 28.3 \\
	TensorMask~\cite{tensormask} & ResNet-101-FPN & 72 & \checkmark & 37.1 & 59.3 & 39.4 & 17.1 & 39.1 & 51.6 \\
	YOLACT~\cite{yolact}  & ResNet-101-FPN & 48 &\checkmark &  31.2  & 50.6 & 32.8 & 12.1 & 33.3 & 47.1 \\

	\textbf{PolarMask} & ResNet-101-FPN & 12 & $\circ$ &30.4  &51.9  &31.0  &13.4  & 32.4 &42.8 \\
	\textbf{PolarMask} & ResNet-101-FPN & 24 & \checkmark &32.1  &53.7  &33.1  &14.7  & 33.8 &45.3 \\
	\textbf{PolarMask} & ResNeXt-101-FPN & 12 & $\circ$ &32.9  &55.4 &33.8 &15.5  &35.1  &46.3 \\
	\textbf{PolarMask} & ResNeXt-101-FPN-DCN & 24 & \checkmark &36.2  &59.4 &37.7 &17.8  &37.7  &51.5 \\
\end{tabular}

%% file: tables/table4.tex
\def\x{{$\footnotesize \times$}}
\small 
\begin{tabular}{l|c|c|c|c}
    \hline
	method & backbone & size & GFLOPs & Parameters   \\
	\hline
	FCOS~\cite{FCOS} & R-50 & $800\times1280$ & 200.4 & 32.02M  \\
    PolarMask w/o box & R-50 & $800\times1280$ & 202.3 & 32.09M  \\
    PolarMask w/ box & R-50 & $800\times1280$ & 252.7 & 34.46M  \\
    YOLACT~\cite{yolact}  & R-50 & $800\times1280$ & 196.8 & 31.16M  \\
    Mask R-CNN~\cite{maskrcnn} & R-50 & $800\times1280$ & 275.5 & 44.18M  \\
    \hline
\end{tabular}

%% file: tables/table5.tex
\def\x{{$\footnotesize \times$}}
\small 
\begin{tabular}{l|l|c|c|c}
    \hline
	method & backbone  &Device & FPS & Time(ms)  \\
	\hline
    PolarMask & R-101  & V100 & \textbf{12.3} & 81  \\
    PolarMask & R-101-DCN & V100 & 8.18 & 122  \\
    TensorMask~\cite{tensormask}  & R-101 & V100 &2.63 & 380  \\
    Mask R-CNN~\cite{maskrcnn} & R-101 & M40 & 5.12 & 195  \\
    \hline
\end{tabular}

%% file: tables/table3.tex
\def\x{{$\footnotesize \times$}}
\small 
\begin{tabular}{l|l|cc|c}
    \hline
	method & backbone & epochs & ms-train  & AP  \\
	\hline
    PolarMask & R-50 & 12 & $\circ$ & 29.1 \\
     & R-50-DCN & 12 & $\circ$ & 32.0 \\
     & R-50 & 24 & \checkmark & 30.5 \\
     & R-50-DCN & 24 & \checkmark & 33.3 \\
     \hline
     & R-101 & 12 & $\circ$ & 30.4 \\
     & R-101-DCN & 12 & $\circ$ & 33.5 \\
     & R-101 & 24 & \checkmark & 31.9 \\
     & R-101-DCN & 24 & \checkmark & 34.3 \\
     \hline
     & X-101 & 12 & $\circ$ & 32.6 \\
     & X-101-DCN & 12 & $\circ$ & 34.9 \\
     & X-101 & 24 & \checkmark & 33.5 \\
     & X-101-DCN & 24 & \checkmark & 35.9 \\
     \hline
\end{tabular}

%% file: tables/table6.tex
\def\x{{$\footnotesize \times$}}
\small 
\begin{tabular}{l|c|c|c|c}
    \hline
	method & modules  &AP & AP$_{50}$ & AP$_{75}$  \\
	\hline
    ESE-Seg    & $\circ$  & 21.6 & 48.7 & 22.4  \\
    PolarMask  & $\circ$  & 25.1 & 46.4 & 24.5  \\
    PolarMask  & $\checkmark$  & 29.1 & 49.5 & 29.7  \\
    \hline
\end{tabular}